\documentclass[runningheads]{llncs}
\usepackage{graphicx}

\usepackage{tikz}
\usepackage{comment}
\usepackage{amsmath,amssymb} 
\usepackage{color}

\usepackage[accsupp]{axessibility}  

\usepackage[width=122mm,left=12mm,paperwidth=146mm,height=193mm,top=12mm,paperheight=217mm]{geometry}

\begin{document}
\pagestyle{headings}
\mainmatter
\def\ECCVSubNumber{1733}  

\title{GradientSurf: Gradient-Domain Neural Surface Reconstruction from RGB Video}



\titlerunning{GradientSurf}
%
\author{Crane He Chen\inst{1} \and
Joerg Liebelt\inst{2}}
%
%
\institute{Johns Hopkins University \and
Apple Inc.}

\maketitle

\begin{abstract}
This paper proposes GradientSurf, a novel algorithm for real time surface reconstruction from monocular RGB video. 
Inspired by Poisson Surface Reconstruction, the proposed method builds on the tight coupling between surface, volume, and oriented point cloud and solves the reconstruction problem in gradient-domain. 
Unlike Poisson Surface Reconstruction which finds an offline
solution to the Poisson equation by solving a linear system after the scanning process is finished, our method finds online solutions from partial scans with a neural network incrementally where the Poisson layer is designed to supervise both local and global reconstruction. 
The main challenge that existing methods suffer from when reconstructing from RGB signal is a lack of details in the reconstructed surface. We hypothesize this is due to the spectral bias of neural networks towards learning low frequency geometric features. To address this issue, the reconstruction problem is cast onto gradient domain, where zeroth-order and first-order energies are minimized. The zeroth-order term penalizes location of the surface. The first-order term penalizes the difference between the gradient of reconstructed implicit function and the vector field formulated from oriented point clouds sampled at adaptive local densities. For the task of indoor scene reconstruction, visual and quantitative experimental results show that the proposed method reconstructs surfaces with more details in curved regions and higher fidelity for small objects than previous methods.
\end{abstract}

\section{Introduction}
\begin{figure*}[t]
  \centering
    \includegraphics[width= 0.9\linewidth]{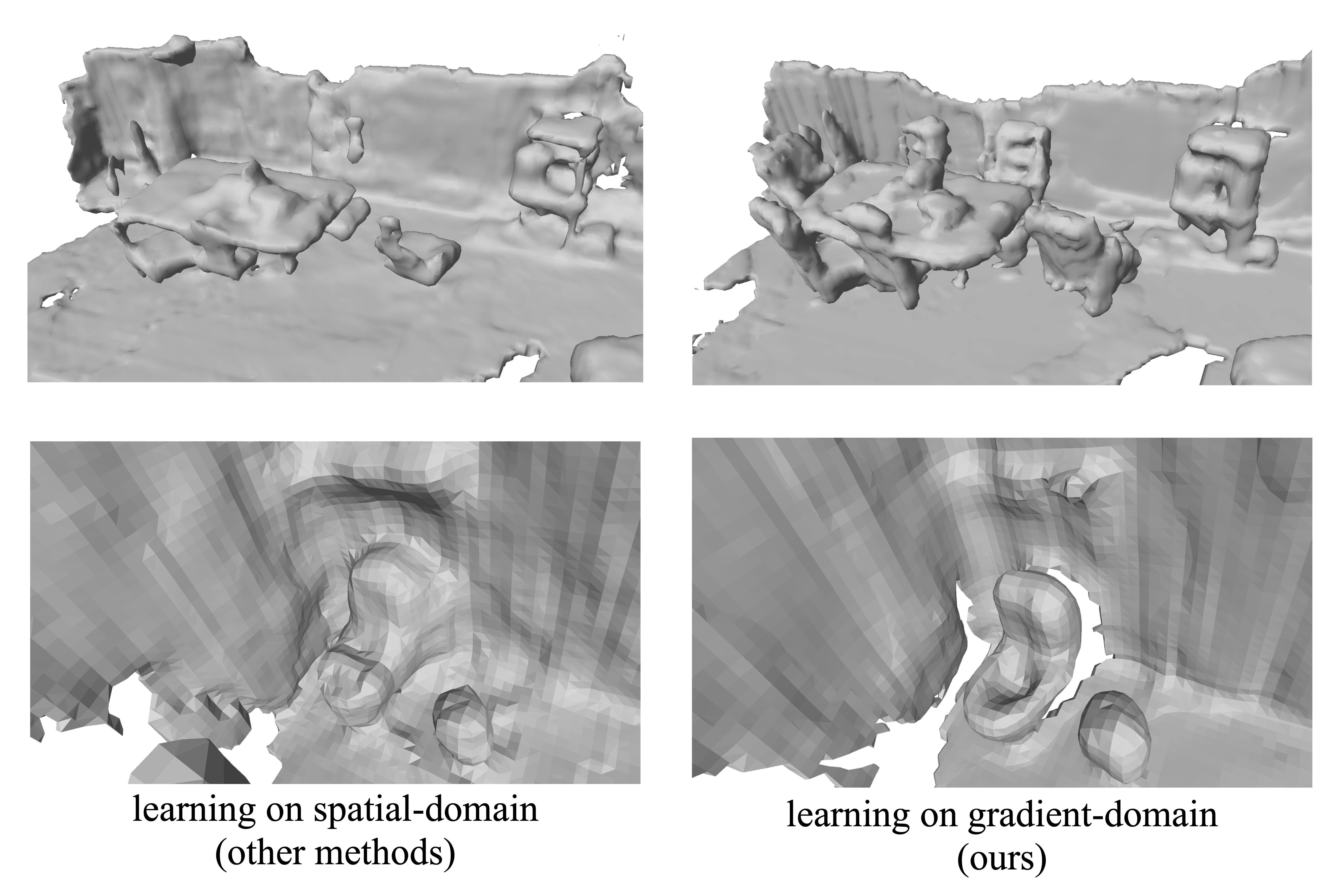}
  \caption{Visualization of real time indoor scene reconstruction from an RGB video with SLAM poses. Existing methods miss smaller objects or detailed curved regions, while
the proposed method preserves more high frequency features by learning from an oriented point cloud.}
  \label{fig:teaser}
\end{figure*}
Surface reconstruction from different sensor signals is a classical problem in 3D computer vision and computer graphics. Well known techniques in this field include Poisson Surface Reconstruction~\cite{kazhdan2006poisson}, volumetric depth fusion~\cite{curless1996volumetric} and kinect fusion~\cite{newcombe2011kinectfusion}. Recent years have seen increasing interest in immersive real time interactive experiences in the field of Augmented Reality, where surface reconstruction plays a key role. Assuming that accurate dense depth information is available and runtime is not a concern, the aforementioned methods achieve impressive performance. However, hardware constraints in these AR scenarios result in limited access to depth information. Moreover, running real time is of crucial importance for the experience to be interactive. This brings us to the challenging problem of reconstructing surfaces in real time on devices without active depth sensors. A neural network is a natural solution for this problem, because it 1) frees the inference stage from iteratively optimizing the energy function by relegating the expensive parts of the computation to the training process; 2) has been proven excellent feature extractors for RGB images.

Despite the recent success of data-driven point cloud processing~\cite{qi2017pointnet,qi2017pointnet++,qi2018frustum,qi2020imvotenet,wang2019dynamic} and neural rendering~\cite{mildenhall2020nerf,zhang2020nerf++,li2020crowdsampling,li2021neural,martin2021nerf,wu2021rendering}, the intersection between 3D vision and graphics with machine learning is still at its infancy. Neural networks originally found their success in image processing, but adaptation from handling image to geometric representation for resolving inverse problems is under-explored. Major challenges include what rules should be applied for the supervision, and on which domain should the rules be placed. The simplest and most convenient way is to carry out calculations on spatial domain. We argue that this is not the only (or necessarily the best) option opposed to frequency domain or gradient domain. One example that explores the superiority of frequency domain over spatial domain is NerF~\cite{mildenhall2020nerf}. As shown in Fig. 4 of NerF~\cite{mildenhall2020nerf}, without the positional encoding with sinusoidal basis functions at different frequencies, the performance would experience a severe drop. 

In this work, we explore neural surface reconstruction on gradient-domain, and consider oriented point cloud to be a preferred 3D geometric representation for supervision associated to the domain. The reasons are threefold: First,  orienting the point clouds with surface normals naturally formulates a vector field which indicates roughly the same information as gradient domain derived from differentiating the implicit functions. Second, point clouds provide arguably the simplest representation for shapes. Acquisition of point clouds from other representations such as meshes or depth maps reduces to sampling, which is straightforward. Third, point clouds provide a representation of geometry without imposing any structural constraints. As such, local density can be controlled either by adjusting the sampling strategy or through the design of sensing hardware.

This paper develops a framework that reconstructs surface in absence of depth signal. In order to achieve this, a neural network predicts a truncated implicit function from learned image feature and camera trajectory. Surface is extracted from the implicit function using a non-learnable marching cubes. Supervision of the reconstruction is placed on the gradient-domain. In the context of indoor scene reconstruction, truncation is essential, so that only the grid cells around the surface participate in the back propagation, whereas cells far from the surface should not impact the loss function. More specifically, the optimization is performed using the oriented points for supervision. Similar to a multigrid solver, our system is hierarchical and the optimization is carried out both locally and globally -- training a local CNN and a global RNN architecture in one pipeline. In this way, although the surface is reconstructed incrementally in a local manner, the reconstruction is consistent across the scene. To better aggregate local geometric feature for training, we consider different ways in which point clouds contribute to the loss function when performing supervision, including Poisson-disk sampling and curvature-based sampling. Compared to previous methods such as~\cite{murez2020atlas,sun2021neuralrecon}, the proposed method demonstrates more detail in curved regions and higher fidelity for small objects. See Fig.~\ref{fig:teaser} for examples. 

\section{Related Works}
A vast amount of literature exist in this field, yet only a subset of them are mentioned here due to the limitation of pages. Refer to survey papers~\cite{berger2017survey,herbort2011introduction} for more complete reviews.
\subsection{Surface Reconstruction}
The challenging problem of surface reconstruction has been addressed using both explicit and implicit approaches. Unlike explicit reconstruction where meshes are obtained by manipulating topological data structures such as incidence matrices or half edges to interpolate and connect the point sets, implicit reconstruction fits implicit functions on regular grids to the points sets. Implicit reconstruction is more robust to sensor noise, and its output is guaranteed to be a manifold. 

One pioneering implicit approach~\cite{hoppe1992surface} reconstructs a surface from an unorganized point set by fitting an oriented tangent plane to each point. To better understand the impact of noise, local curvature, and sampling density on surface normal estimation, \cite{mitra2003estimating} carried out an analysis with local least square fitting. Aforementioned methods solve the reconstruction problem or estimate surface normal by dealing with local geometric features. An alternative is to handle the problem globally. Radial basis functions (RBFs) offers a compact functional description for implicit surface~\cite{carr2001reconstruction}. More closely related to our work is Poisson Surface Reconstructions. A simple and elegant Poisson problem was formulated in~\cite{kazhdan2006poisson} based on the observation that the gradient of the implicit function (indicator function) matches the vector field formulated by the oriented point cloud. To obtain sharper details, point interpolation was incorporated within the Poisson problem using a screening term~\cite{kazhdan2013screened} with the insight that surface should pass near the input samples. Envelope constraints~\cite{kazhdan2020poisson} were introduced by adapting the basis functions used for discretization to support targeted Dirichlet constraints. 

Rather than writing deterministic algorithms that make geometric inferences, an alternative is to train a neural network to fit a model to a dataset. In this case, combination of insights between geometry processing and learning pipeline can be powerful. Point2Mesh~\cite{hanocka2020point2mesh} optimizes a neural network to shrink-wrap a point cloud from convex hull into a surface with fine details. AtlasNet~\cite{groueix2018papier} designs an encoder-decoder architecture to reconstruct a surface by learning the deformation of parametric surface elements such as a set of squares or a sphere. ~\cite{morreale2021neural} shows that the neural network of ~\cite{groueix2018papier} goes beyond reconstruction task by introducing a novel representation of 3D data named \textit{neural surface}, and applied the network as a parametric representation of both individual surfaces as well as inter-surface maps. 

\subsection{3D Capture from Videos}
Low latency is essential for interactive applications which perform real-time inference on mobile devices. Thus, the ability to reconstruct incrementally from a sequence of data is appealing. 

Existing algorithms work well on incremental surface reconstructions when accurate depth maps are available ~\cite{curless1996volumetric,newcombe2011kinectfusion,niessner2013real,dai2017bundlefusion,azinovic2021neural,schertler2017field}. In practical AR/VR scenarios, access to depth information is limited. Structure from motion methods (SFM)~\cite{snavely2006photo,agarwal2011building,wu2013towards} can reconstruct geometry from RGB videos, but the reconstructed scenes are typically point clouds rather than a surface. Considering single RGB image as input, ~\cite{dahnert2021panoptic} proposes a panoptic 3D scene reconstruction that unifies geometric reconstruction with segmentation from a single shot. However, it is not making use of the coherence from temporal scale.

When considering RGB videos as input, \cite{murez2020atlas} pioneered one solution by proposing a CNN-based machine learning pipeline, where the network takes an RGB video as well as per-frame SLAM camera pose as inputs.  The visual hull at the union of several RGB frames can be derived from the intrinsic parameters and trajectory of the RGB camera. Attaching a local regular grid to this visual hull, the learned 2D image features are back projected into a 3D feature volume by filling values into the desired voxel grid. Information from multiple views are then aggregated by averaging. ~\cite{sun2021neuralrecon} makes improvements in temporal coherence by incorporating recurrent neural network (RNN).~\cite{bozic2021transformerfusion} takes a step further on temporal scale by enabling key frame selection at interactive rates with a Transformer. ~\cite{stier2021vortx} incorporates Transformer from a different perspective aiming at reducing irrelevant information by projective occupancy. Aforementioned methods make improvements on temporal scale, but artifacts and lack-of-details still exist spatially. To take a step back, the aggregation of spatial information is still sub-optimal and detail-driven improvements are under-explored.

\subsection{Gradient-domain Processing}
Gradient-domain processing is a classic in computer vision and graphics~\cite{simchony1990direct}. Applications that it can resolve include image stitching~\cite{perez2003poisson,agarwala2004interactive,levin2004seamless,kazhdan2008streaming}, texture stitching~\cite{prada2018gradient}, smoothing/sharpening~\cite{bhat2008fourier,bhat2010gradientshop,chuang2016gradient}, surface reconstruction~\cite{kazhdan2006poisson}, etc. Operations are defined on gradient-domain using the differences between neighboring pixels or simplex. The goal of gradient-domain processing is to reconstruct new images or geometries by integrating the gradient, which boils down to solving a Poisson's equation. Multiple ways can be used to find solutions of Poisson's equation, including FFT, linear system, and leveraging neural networks. 

\section{Method}
\begin{figure*}[t]
  \centering
    \includegraphics[width= 0.95\linewidth]{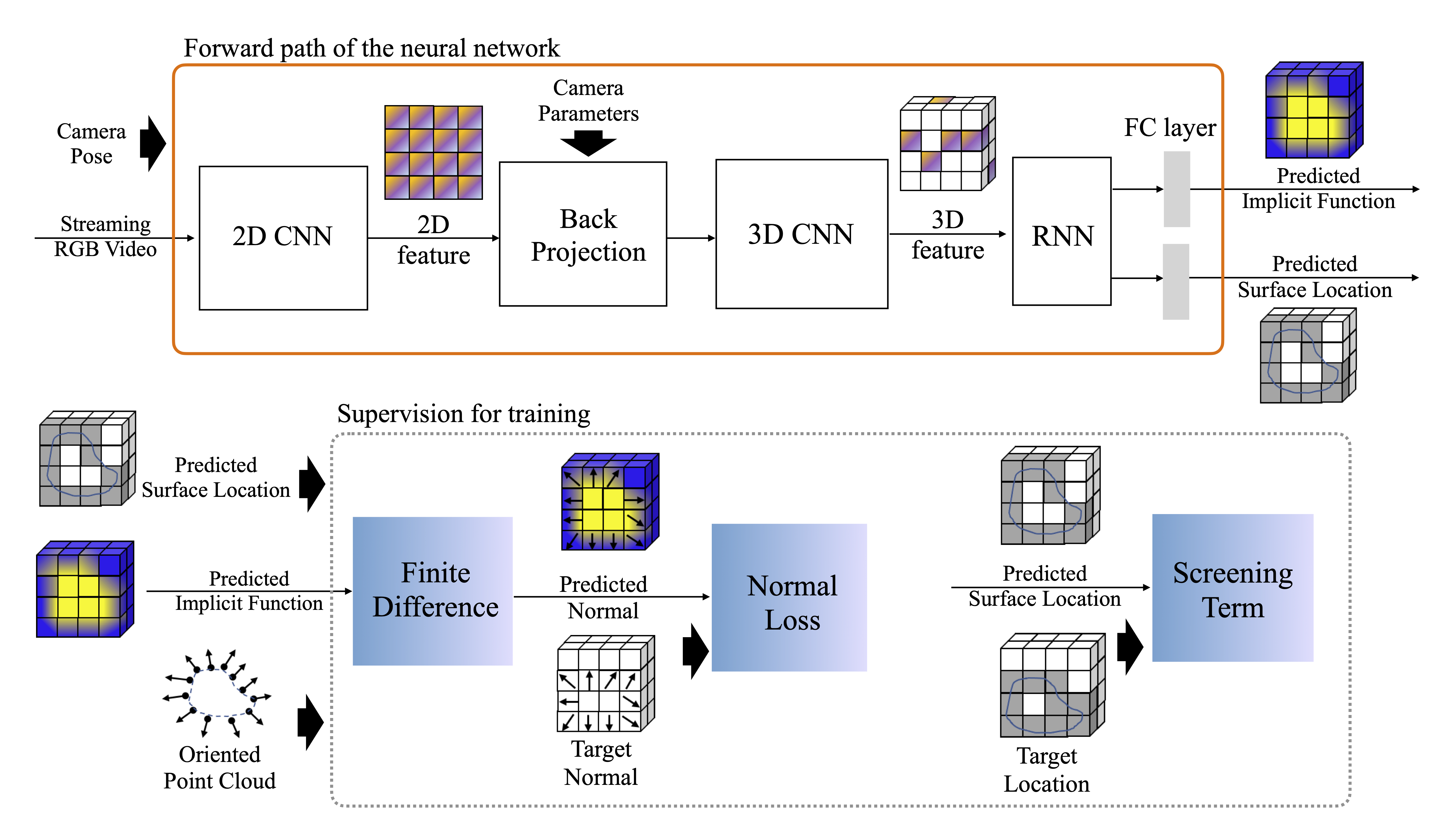}
  \caption{Overview of our method. The method takes a streaming RGB video as the input signal. A 2D CNN backbone generates 2D feature map from the input frames. Using camera parameters pre-obtained by SLAM, the feature map is back projected into 3D camera frustum. 3D features goes through a 3D CNN backbone and a RNN block that fuses local 3D feature into global 3D feature. In order to train the aforementioned forward path, oriented point clouds are used as reference signals that provide supervision for both normal and location. Predicted implicit function is fed into the finite difference block to predict normals and compare against the normal of oriented PCD. Predicted implicit function is encouraged to be close to zero at GT locations by the screening term.}
  \label{fig:pipeline}
\end{figure*}

We propose a method for real-time surface reconstruction from RGB video by optimizing the weights of a CNN locally and an RNN globally. In our formulation, a neural network is used to reconstruct the surface by computing the indicator function that akin to the solution of a screened Poisson equation~\cite{kazhdan2013screened}, where an oriented point cloud is taken as the reference signal during training. Marching cubes is used to extract the reconstructed surface from that implicit representation. The implicit function is calculated from coarse to fine using a multi-resolution neural network. In this section, we first describe the problem formulation, then discuss three critical components of our method. An overview of the method is shown in Fig.~\ref{fig:pipeline}.

\label{sec:method}
\subsection{Problem Formulation}
Given a monocular RGB sequence $V = \{f_1, f_2, ..., f_n\}$ with corresponding camera intrinsics $K_n$ and poses $T_n$, our goal is to learn a back-projection from pixel to voxel 
\begin{equation}
  (X,Y,Z) = \phi(u_1,v_1, u_2, v_2, ..., u_n, v_n).
  \label{eq:the learning problem}
\end{equation}
Unlike most learning-based methods minimizing a loss function defined in terms of a reconstructed volumetric function on spatial-domain, we use the neural network to predict a truncated indicator function that is akin to the solution of a screened Poisson equation~\cite{kazhdan2013screened}.  

We use a neural network to encode RGB video $V$ into a 3D implicit function $\chi$ that contains geometric information, 
\begin{equation}
  \chi = \psi_{\theta} (f_1, f_2, ... f_n).
  \label{eq:network}
\end{equation}
The oriented point cloud is taken as the reference signal during training. We seek an implicit function $\chi$ that minimizes the following energy as the loss function
\begin{equation}
  E = w_0 \sum_{p\in P}\|\chi(p) \| + w_1 \sum_{(q,n)\in Q}\|\nabla \chi(q) - n \|,
  \label{eq:optimization problem}
\end{equation}
where $P$ is a subset of points used to evaluate the location constraint of the loss and $Q$ is a subset of oriented points used to evaluate the gradient constraint of the loss. It should be noted that in the context of indoor scene reconstruction, $\chi$ needs to be truncated to prevent any point sampled far away from the reconstructed region to have any impact on the energy. This is because when incrementally reconstructing a room scene, local geometric features are the main cue for reconstruction. The truncation step is optional if the goal is to reconstruct a sculpture or a solid object, since in those scenarios it can be beneficial to globally smooth the reconstruction in order to fill in areas of partial scanning and missing information. This does not apply to indoor scene reconstruction where global smoothing and hallucinating to infill unscanned unknown areas is detrimental. In this paper, reconstructing indoor scenes is our main goal, and $\chi$ is always truncated.

Unlike conventional reconstruction algorithms where the same signal from a sensor is used for both generating the implicit function and the target vector field, in our formulation, the oriented point cloud only serves as the target signal during training, whereas during inference the point cloud is not required. 

\subsection{The Proposed Layer}
A coarse-to-fine neural network is adopted to learn the truncated implicit function. The topology of the network resembles that in~\cite{sun2021neuralrecon}. We propose to add a Poisson layer that enables supervision on gradient-domain.
Observing that the normals of oriented points should agree with the normals of the reconstructed surface, we aim to characterize this correspondence within the loss function. A naive approach would be to extract the iso-surface and use its normals. Unfortunately, this would not fit into a training framework because there is no appropriate differentiable extraction process that doesn't bring in additional artifacts, and switching unnecessary conversions between implicit and explicit representation of the shape would be inefficient. Instead, we use the observation that the gradients of the truncated indicator function agree with the normals of the oriented points and use finite differences to estimate the gradient at a voxel -- an operation that is differentiable. For example, using central differences, we have:
\begin{align}
  N_x &= \frac{\partial \chi(i,j,k)}{\partial x} \approx \frac{\chi(i+1, j, k) - \chi(i-1, j, k)}{2v},
  \label{eq:finite diff1}\\
  N_y &= \frac{\partial \chi(i,j,k)}{\partial y} \approx \frac{\chi(i, j+1, k) - \chi(i, j-1, k)}{2v},
  \label{eq:finite diff2}\\
  N_z &= \frac{\partial \chi(i,j,k)}{\partial y} \approx \frac{\chi(i, j, k+1) - \chi(i, j, k-1)}{2v}.
  \label{eq:finite diff3}
\end{align}
where $v$ is the voxel with.

To  be able to compare gradients, which are associated with voxels, with oriented point clouds, we discretize the oriented points by tri-linearly splatting them into the voxel grid and storing the sum of the splatted normals as well as the the sum of the splatting weights within each voxel.

\subsection{Spectral Bias Handling}
Key to our method is taking the spectral bias into consideration when designing the framework. Neural networks tend to prioritize learning lower frequency functions~\cite{rahaman2019spectral}. This could lead to a tendency of under-performance for the representation of high frequency variations. In perception-oriented problems such as image classification, object detection, and semantic segmentation, this bias is less an issue because the high frequency features usually represent intra-class difference which we want to eliminate in order to classify individuals with minor difference into the same class, while inter-class difference is usually represented by low frequency feature. However, when it comes to inverse problems such as rendering and reconstruction where geometry plays an important role and higher fidelity at details is the goal, this spectral bias towards low frequency leads to hallucination and over-smoothing. 

To address the spectral bias, larger weights need to be placed on high frequency components. NerF~\cite{martin2021nerf} and SIREN~\cite{sitzmann2020implicit} adopt the strategy of positional encoding for this purpose. The idea is to use encoding function in the format of sinusoidal basis at different frequencies to composite the coordinates. 

Different from those works, we leverage the Laplacian for spectral analysis to highlight high frequency signals. As observed by Bhat~{\em et al.}~\cite{bhat2008fourier}, when solving the optimization problem in the gradient-domain, the loss function gives more importance to the preservation of high-frequency details. 

\subsection{Sampling Strategies}
Additionally, to address the spectral bias problem, geometrically, point clouds are sampled in an adaptive manner. 
In particular, we aggregate the local geometric feature by calculating the weighted average of points locating in the same hyper cube.  The small clusters of points are  splatted into voxel grids by a weighted sum, where the weight is the distance between a point and the cell center. It should be noted that in our experimental setting, it is assumed that geometry of the training set is pre-known. Oriented point clouds for supervision are sampled directly from the ground truth meshes.

We find that the oriented point samples contribute differently to the interpolation and gradient-fitting components of the loss function in Eq.~(\ref{eq:optimization problem}). Specifically, we find that while the interpolation component is more effective when samples are uniformly distributed, the gradient-fitting component is more effective when the samples come from high-curvature regions. 

We use Poisson disk sampling for the interpolation loss (point cloud $P$), and the local surface curvature guided sampling for the gradient-fitting loss (point cloud $Q$). 

Poisson disk sampling generates points that are evenly distributed, with any two points no closer to each other than a threshold. This sampling strategy is ideal for the screening term because it roughly produces a dense uniformly distributed point cloud. We leveraged the Poisson disk sampling tool from Meshlab.

To make fundamental improvements in curved regions and details, we use curvature to guide the sampling process. Curvature determines local geometry of a surface. Curvature estimation closely relates to the eigen-decomposition of shape operator. The estimation of shaper operator introduces errors. In the context of detail-preserving surface reconstruction, the luxury of knowing exact values of curvatures among all directions is not necessary. Instead, total curvature is already sufficient in representing all the highly curved regions. In~\cite{10.1145/3587421.3595439}, the author observed that a simpler and more intuitive way to estimate curvature is by calculating the Dirichlet energy of Gauss map. This method is mathematically rigorous, but relatively slower to run on a dataset with thousands of large meshes. Therefore, it could be less suitable for data-driven settings for real-world scenes. However, we can borrow the concept from the paper, and speed it up by a reasonable simplification:

For each triangle on a mesh, we estimate its curvature using vertex normals, edge lengths, and face area using
\begin{equation}
 \mathcal{C} =\left (\left \|\frac{N_A - N_B}{\mathcal{E_{AB}}}\right\|^2 + \left\|\frac{N_B - N_C}{\mathcal{E_{BC}}}\right\|^2 + \left\|\frac{N_A - N_C}{\mathcal{E_{AC}}}\right\|^2 \right)\cdot \mathcal{A},
  \label{eq:local surface curvature sampling}
\end{equation}
where $N_A$, $N_B$, $N_C$ represent the normals at the vertices of triangle $ABC$, $\mathcal{E_{AB}}$, $\mathcal{E_{BC}}$, $\mathcal{E_{AC}}$ are the lengths of the edges, and $\mathcal{A}$ is the area of the triangle. This descriptor measures the variation for the normal across a triangle.

\begin{figure}[h]
  \centering
    \includegraphics[width= 0.3\linewidth]{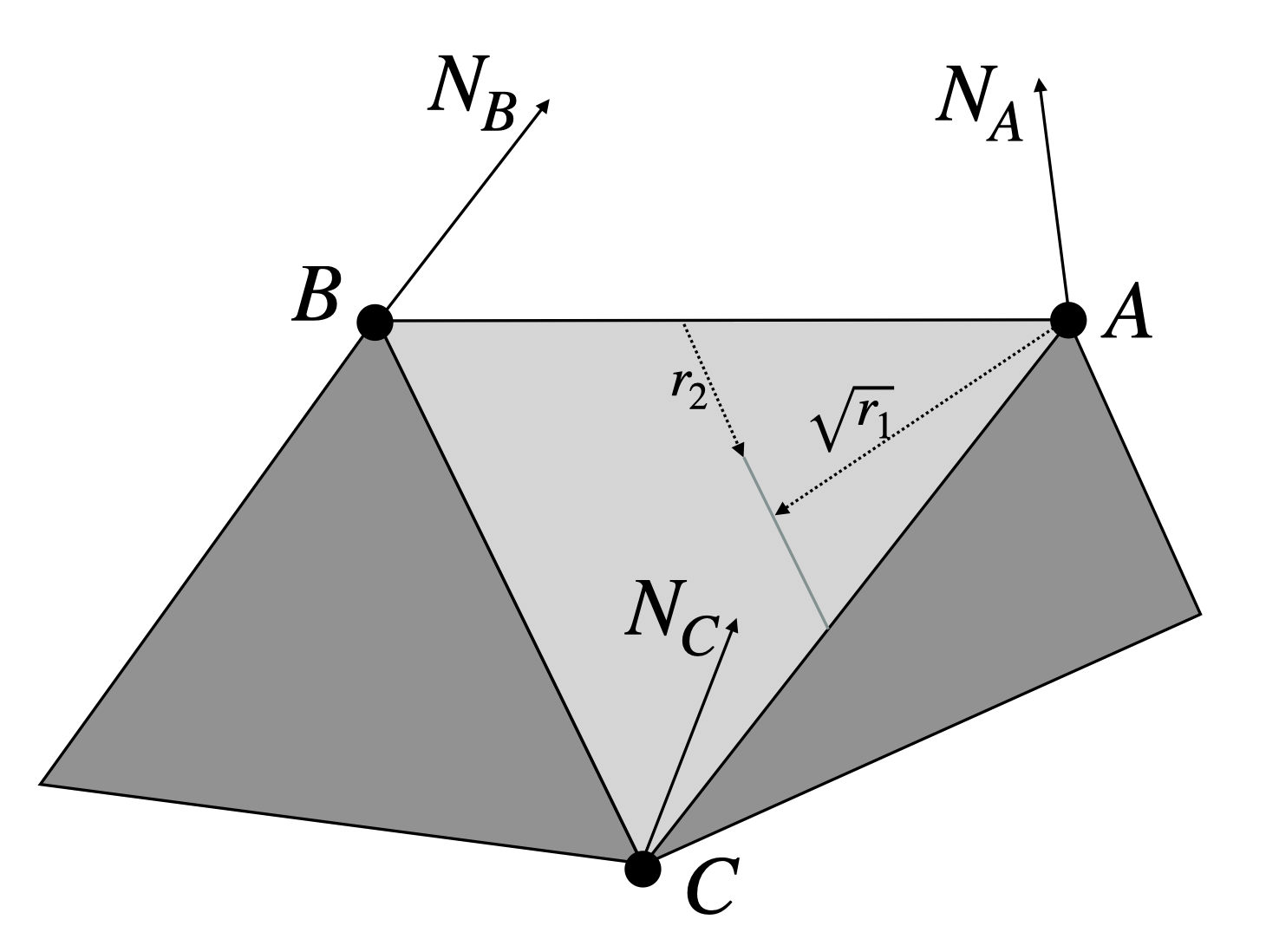}
  \caption{Illustration of local surface curvature guided point sampling}
  \label{fig:local curvature}
\end{figure}

The curvature-based sampling is developed in a manner similar to the triangle area guided sampling~\cite{osada2002shape}. The sampling strategy, as illustrated in Fig.~\ref{fig:local curvature}, follows these steps: First, for each triangle, we store its curvature in an array along with the cumulative curvature of triangles visited so far. Second, we select a triangle with the probability proportional to its local curvature by choosing a random number between 0 and the total accumulated curvature, followed by a binary search. Finally, for each selected triangle, two random numbers $r_1, r_2 \in [0,1]$ are selected to find the location and normal of the selected point within the triangle
\begin{equation}
 P = (1 - \sqrt{r_1})A + \sqrt{r_1}(1 - r_2)B + \sqrt{r_1}r_2C,
  \label{eq:local surface curvature sampling}
\end{equation}
\begin{equation}
 N = (1 - \sqrt{r_1})N_A + \sqrt{r_1}(1 - r_2)N_B + \sqrt{r_1}r_2N_C.
  \label{eq:local surface curvature sampling}
\end{equation}

\begin{figure*}[!hbt]
  \centering
    \includegraphics[width= 1.0\linewidth]{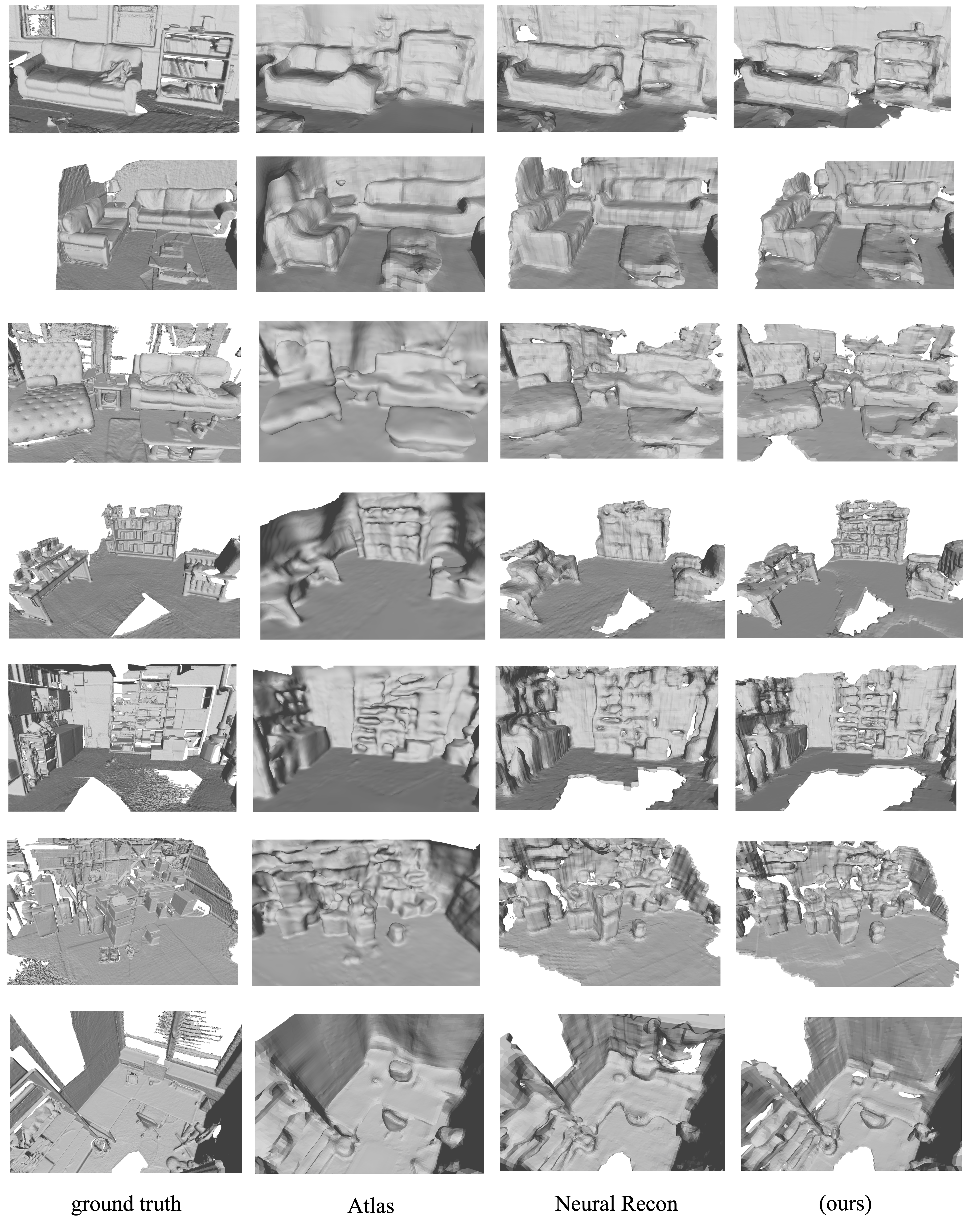}
  \caption{Qualitative comparison of scene reconstruction with baselines. As can be observed that Atlas tends to be more complete while at the same time over-smoothing and deviating from ground truth, especially for windows or the middle of floor were the user stands and which were never scanned. Neural Recon tends to lose details for small objects and edge details. Our method produces sharper results, has higher fidelity for edges, smaller objects, and curved objects.}
  \label{fig:comp}
\end{figure*}
\section{Experiments}
In this section we validate both visually and quantitatively how our algorithm improves the quality of real-time reconstruction for indoor scenes. An overview of the pipeline we use to evaluate the proposed method is illustrated in Fig.~\ref{fig:pipeline}. We train the network at three different voxel resolutions, 4cm, 8cm, 16cm. All results we show in this section are trained from scratch. We use ScanNet dataset~\cite{dai2017scannet} which provides real scanned indoor scenes with rich annotations to validate our method. The dataset provides RGB-D sequences and camera trajectories associated with them. The depth sensor has an accuracy and quality similar to that of Kinect v1. Note that, as with most real-world data, reconstruction at very high resolution might not be desirable as it can reproduce noise like scanner error and camera shake. In our experimental setting, ground truth geometry is also considered as pre-known for training and evaluation. For testing, only RGB sequence and camera poses are required as input. Train/val/test split of the dataset follows that of ~\cite{murez2020atlas,sun2021neuralrecon}. All the experimental results are trained from scratch, no pre-training is involved.

We further remark that despite the significant effort to formalize reconstruction metrics, there is still work to be done and better values using metrics like the F-score do not necessarily correspond to qualitatively better reconstructions~\cite{hanocka2020point2mesh}. To this end, in addition to the F-score and 2d metrics by back-projecting the reconstructions into the camera, we also present in-depth visualizations.

\subsection{Comparison with Existing Methods}

Both quantitative and qualitative comparisons with existing methods show that the simplicity and efficiency of our method leads to a state-of-the-art performance. Refer to Table~\ref{tab:3D compare} and Table~\ref{tab:2D compare} for quantitative comparison.
\begin{table}[!hbt]
\begin{center}
\caption{3D Metrics Comparison with previous SOTA on ScanNet Dataset}
\label{tab:3D compare}
 \begin{tabular}{ccccl}
\hline\noalign{\smallskip}
Method & Precision & Recall & F-Score & ms/frame\\
\noalign{\smallskip}
\hline
\noalign{\smallskip}
Atlas~\cite{murez2020atlas} & \textbf{0.732} & 0.382 & 0.499 & 459\\
NeuralRecon~\cite{sun2021neuralrecon} & 0.609 & 0.450 & 0.516 & 47\\
ours &0.641 & \textbf{0.455} & \textbf{0.530} & \textbf{45}\\
\hline
\end{tabular}
\end{center}
\end{table}
\begin{table}[!hbt]
\begin{center}
\caption{2D Metrics Comparison with previous SOTA on ScanNet Dataset}
\label{tab:2D compare}
 \begin{tabular}{ccccl}
\hline\noalign{\smallskip}
Method & Abs Rel & Abs Diff & Sq Rel & RMSE\\
    \noalign{\smallskip}
    \hline
    \noalign{\smallskip}
    Atlas~\cite{murez2020atlas} & \textbf{0.065} & 0.123 & 0.045 & 0.251\\
    NeuralRecon~\cite{sun2021neuralrecon} & \textbf{0.065} & \textbf{0.106} & \textbf{0.031} & 0.195\\
    ours & \textbf{0.065} & 0.107 & 0.032 & \textbf{0.121}\\
\hline
\end{tabular}
\end{center}
\end{table}
The proposed algorithm takes oriented point cloud from two sampling strategies as supervision data. Normal supervision takes curvature-based sampled point clouds as reference, while location supervision takes Poisson disk sampled point clouds as reference. Running times are compared in Table~\ref{tab:3D compare}. These are calculated by averaging the incremental reconstruction of one crop of the room scene divided by the number of frames involved. 

The performance improvement can be observed by comparisons shown in Fig.~\ref{fig:comp}. Our method has sharper results with better details. The edges of shelves are sharper, as can be observed in first, fourth, fifth rows of the figure. Small objects are more accurately preserved. One example is on the first row, check out the object on top of book shelf. Curved objects are more precisely reconstructed. One example is the litter bin in the last row.

\begin{figure*}[!hbt]
  \centering
    \includegraphics[width= 0.7\linewidth]{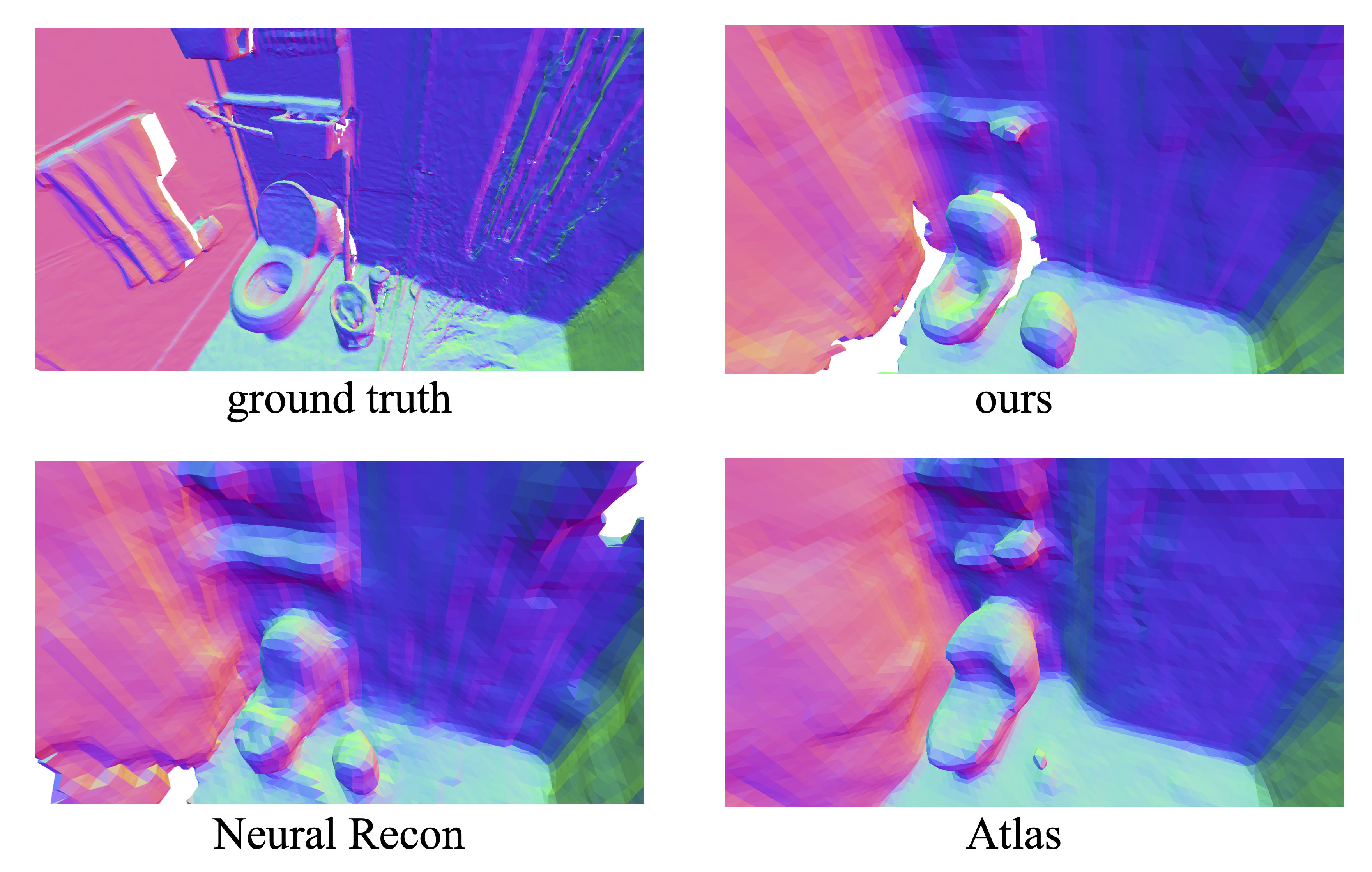}
  \caption{Comparison of normal maps for details. It gives a clear visual intuition with surface normals. Our method does a better job in reconstructing the toilet.}
  \label{fig:normal_map}
\end{figure*}

\begin{figure*}[!hbt]
  \centering
    \includegraphics[width= 1.0\linewidth]{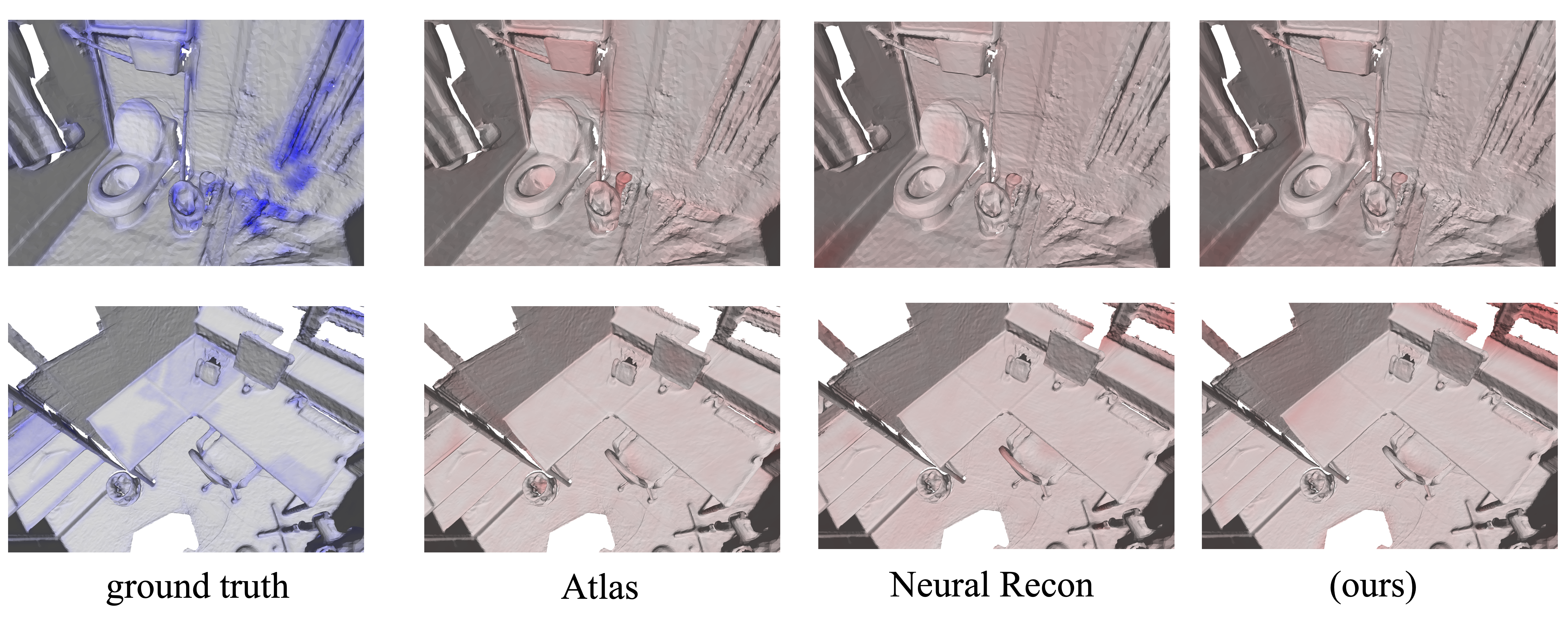}
  \caption{Visualization of reconstruction errors rendered as vertex heat map onto ground truth. Heat map of the left most column represents total curvature of vertices, the more blue, the larger total curvature is. Heat map of the right three columns represent distance between reconstructed surface and ground truth surface, the more red, the larger reconstruction error is.}
  \label{fig:vertexmap}
\end{figure*}

In addition to basic qualitative comparison, we also present a zoomed-in comparison of the normal maps for one object, the toilet, in Fig.~\ref{fig:normal_map}. We used Open3d~\cite{Zhou2018} API to generate the normal maps. The reconstruction result of our method is aesthetically more appealing due to the more precise reconstruction of curved regions. In this example, Neural Recon suffers from incompleteness, whereas Atlas suffers from over-smoothing.
Another visualization is the reconstruction errors rendered as vertex heat maps onto the ground truth mesh, shown in Fig.~\ref{fig:vertexmap}. The left most column presents a heat map representing the value of of total curvature. The higher total curvature, the more blue. Right three columns show heat maps representing distance between predicted surface and ground truth surface. We adopt a simple way to measure the distance, sampling two point clouds by retrieving all the vertices. Then for each point from the source point cloud, we find its k nearest neighbors in the target point cloud. The calculate the sum of Euclidean distance between this point and the cluster of neighboring points. 
\subsection{Ablation Study of Sampling Strategies}
\begin{figure*}[!hbt]
  \centering
    \includegraphics[width= 1.0\linewidth]{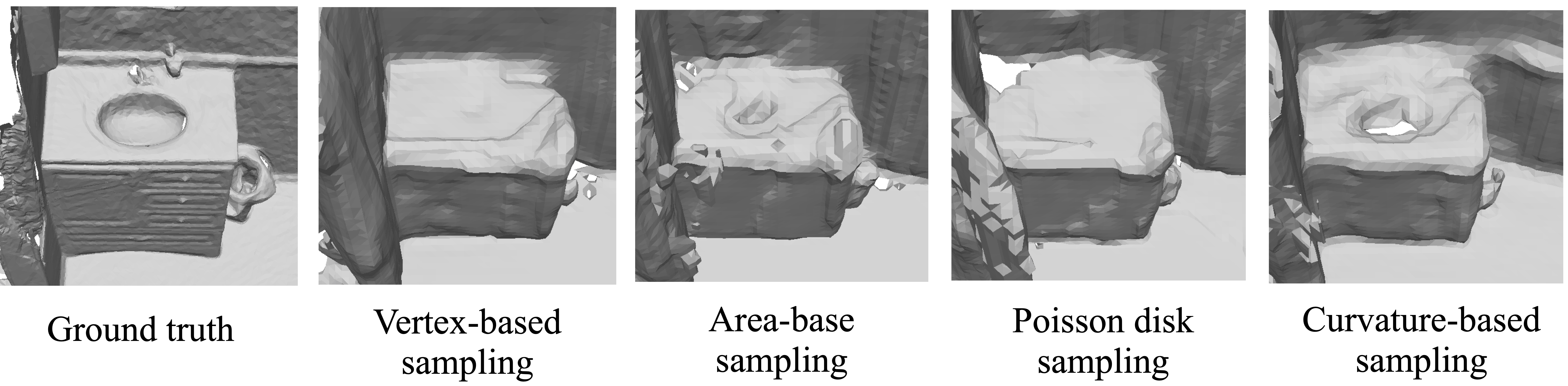}
  \caption{Comparison of the reconstruction result trained with oriented point cloud obtained from different sampling strategies.}
  \label{fig:SamplingAblation}
\end{figure*}

Ablation study is carried out to justify the selection of sampling strategies. Four different sampling strategies, vertex, area-based, Poisson disk, and curvature-based are ablated. For each sampling strategy, we sample one million oriented points from each scene. These points are rasterized into regular grids at the three resolutions. Refer to Fig.~\ref{fig:SamplingAblation} for visual comparison, and Table~\ref{tab:3D ablation} and Table~\ref{tab:2D ablation} for quantitative comparison. Curvature-based sampling performs the best in most metrics, whereas Poisson disk sampling performs the best for precision in 3D. This is due to Poisson disk sampling samples relatively more even than curvature-based, thus has advantage when optimizing the screening term of Eq.~\ref{eq:optimization problem}.
\begin{table}[!hbt]
\begin{center}
\caption{3D Metrics Comparison with Different Sampling Strategies}
\label{tab:3D ablation}
 \begin{tabular}{ccccl}
\hline\noalign{\smallskip}
Sampling Strategy & Precision & Recall & F-Score\\
    \noalign{\smallskip}
    \hline
    \noalign{\smallskip}
    Vertex-based &  0.543 & 0.410 & 0.465\\
    Area-based &  0.529 & 0.392 & 0.449 \\
    Poisson disk & \textbf{0.579} & 0.414 & 0.479\\
    Curvature-based & 0.563 & \textbf{0.439} & \textbf{0.491}\\
\hline
\end{tabular}
\end{center}
\end{table}
\begin{table}[!hbt]
\begin{center}
\caption{2D Metrics Comparison with Different Sampling Strategies}
\label{tab:2D ablation}
 \begin{tabular}{ccccl}
\hline\noalign{\smallskip}
Sampling Strategy & Abs Rel & Abs Diff & Sq Rel & RMSE\\
    \noalign{\smallskip}
    \hline
    \noalign{\smallskip}
    Vertex-based & 0.09 & 0.156 & 0.051 & 0.276\\
    Area-based &  0.093 & 0.162 & 0.053 & 0.283\\
    Poisson disk & 0.084 & 0.142 & 0.049 & 0.253\\
    Curvature-based & \textbf{0.079} & \textbf{0.138} & \textbf{0.041} & \textbf{0.253}\\
\hline
\end{tabular}
\end{center}
\end{table}

\begin{figure*}[!htb]
  \centering
    \includegraphics[width=0.8\linewidth]{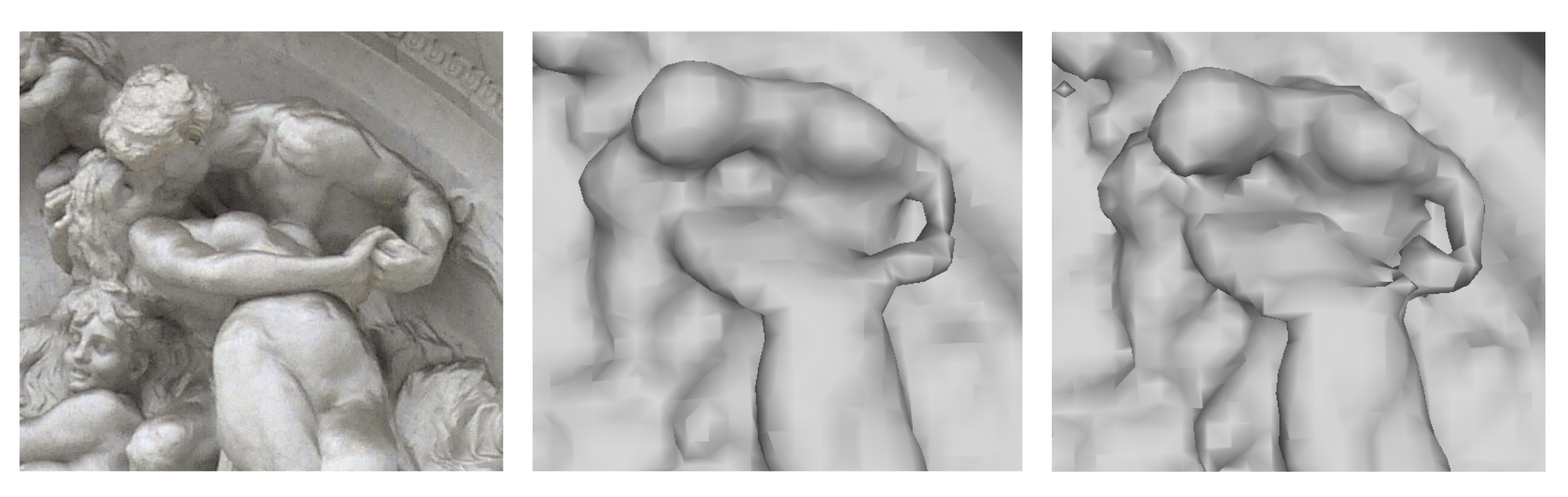}
  \caption{Comparison of details for reconstructed Palacio de Bellas Artes. Depth of tree chosen as eight. Left: RGB photo, Middle: PoissonRecon without emphasizing highly curved regions, Right: PoissonRecon emphasizing highly curved regions.}
  \label{fig:reconstruction_palace}
\end{figure*}

We further observe that giving more attention to highly curved regions help improve thin features and details in surface reconstruction in general. This is not limited to data-driven methods. This is a balancing problem between smoothness prior and data prior. Figure~\ref{fig:reconstruction_palace} shows a comparison of results of details from the reconstruction of the Palacio de Bellas Artes with curvature weighting. The point cloud data was captured by FARO scanner. When reconstructing at the same resolution, emphasizing highly curved regions during reconstruction reproduces sharper details. For example,the detailed muscles are more precisely reconstructed. Since a ground truth mesh is unavailable, we provide a 2D RGB photo for reference.

\subsection{Generalization}
\begin{figure*}[!htb]
  \centering
    \includegraphics[width=0.9\linewidth]{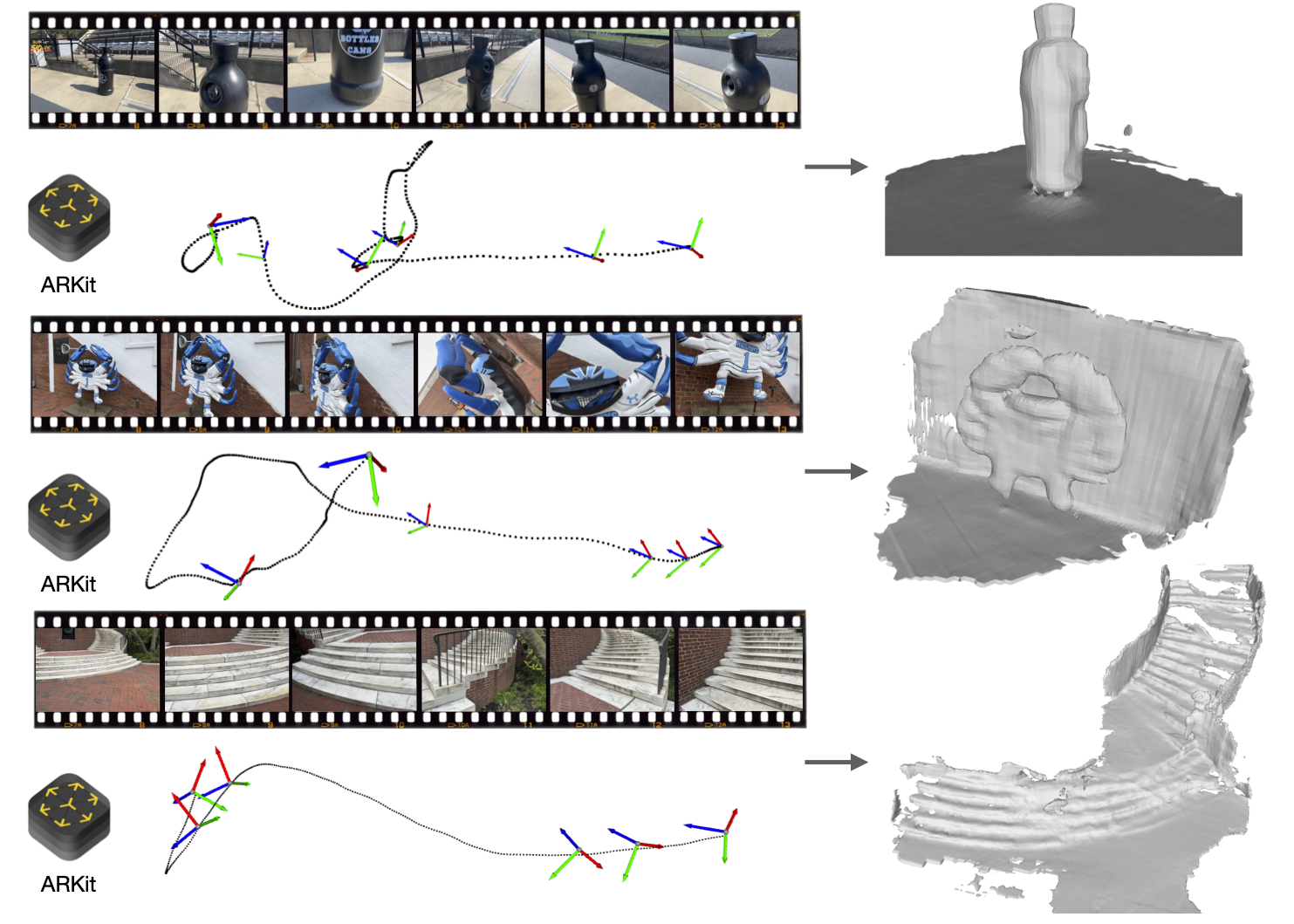}
  \caption{Reconstruction from RGB video captured by iPhone 13 Pro and camera trajectory logged from ARKit. Videos are captured at Homewood campus of the Johns Hopkins University. No LiDAR data was used in the reconstruction process. Reconstruction is performed 33 key frames per second.}
  \label{fig:demo}
\end{figure*}

The trained model generalizes to unseen outdoor RGB videos captured by mobile devices, although the model is trained only 1513 indoor scenes from ScanNet dataset. Several examples are shown in Fig.~\ref{fig:demo}. The videos are collected by iPhone13 Pro, SLAM camera poses are provided by ARKit~\cite{arkit}.

\section{Conclusions}
We present a novel framework that learns to incrementally reconstruct surfaces from RGB videos. Our method is simple, efficient, and geometrically explainable. Local and global geometric features are effectively aggregated on the gradient domain so that the reconstructed surface has sharper details, higher fidelity, and does not suffer from over-smoothing. Although we validate the proposed method with one network architecture, it is not architecture-specific and can be used in conjunction with other architectures such as transformers.

\section{Acknowledgement}
We are grateful to Misha Kazhdan, Ming Chuang, Feng Tang for illuminating discussions, and to Yikang Liao, Kota Hara for their valuable feedback during team standup meetings.


\par\vfill\par

\clearpage
%
%
\bibliographystyle{splncs04}
\bibliography{main}
\end{document}